\documentclass[letterpaper]{article} 
\usepackage{aaai23}  
\usepackage{times}  
\usepackage{helvet}  
\usepackage{courier}  
\usepackage[hyphens]{url}  
\usepackage{graphicx} 
\usepackage{natbib}  
\usepackage{caption} 
\usepackage{algorithm}
\usepackage{algorithmic}

\usepackage{tabularx}
\usepackage{booktabs}
\usepackage{multirow}
\usepackage{array}
\usepackage{xcolor}

\usepackage[utf8]{inputenc} 
\usepackage[T1]{fontenc}    
\usepackage{hyperref}       
\usepackage{url}            
\usepackage{booktabs}       
\usepackage{amsfonts}       
\usepackage{nicefrac}       
\usepackage{microtype}      
\usepackage{graphicx}
\usepackage{amsmath,amssymb} 
\usepackage{color}
\usepackage{epsfig}
\usepackage{tabularx}
\usepackage{booktabs}
\usepackage{multirow}
\usepackage{array}
\usepackage{xspace}
\usepackage{balance}
\usepackage{enumitem}
\usepackage{mathtools}
\usepackage{rotating}
\usepackage{tabulary}
\usepackage[export]{adjustbox}
\usepackage{ctable}
\usepackage{diagbox}
\usepackage{subcaption}
\usepackage{makecell}
\usepackage{bbm}
\usepackage{wrapfig}
\usepackage{newfloat}
\usepackage{listings}
\DeclareCaptionStyle{ruled}{labelfont=normalfont,labelsep=colon,strut=off} 
\floatstyle{ruled}
\newfloat{listing}{tb}{lst}{}
\floatname{listing}{Listing}
\usepackage{amsfonts}
\usepackage{amsmath}

\title{Alignment-Enriched Tuning for Patch-Level Pre-trained Document Image Models}
\author{
    Lei Wang\textsuperscript{\rm 1},
    Jiabang He\textsuperscript{\rm 2},
    Xing Xu\textsuperscript{\rm 2}\thanks{Corresponding Author},
    Ning Liu\textsuperscript{\rm 3},
    Hui Liu\textsuperscript{\rm 4}
}
\affiliations{
    \textsuperscript{\rm 1} Singapore Management University \\
    \textsuperscript{\rm 2} University of Electronic Science and Technology of China \\
    \textsuperscript{\rm 3} Beijing Forestry University \,\,\,
    \textsuperscript{\rm 4} Beijing Rongda Technology Co., Ltd. \\

    lei.wang.2019@phdcs.smu.edu.sg, JiaBangH@outlook.com, xing.xu@uestc.edu.cn \\ liuning0928@bjfu.edu.cn, ryuki122382@gmail.com
}

\usepackage{bibentry}

\begin{document}

\maketitle

\begin{abstract}
Alignment between image and text has shown promising improvements on patch-level pre-trained document image models. 
However, investigating more effective or finer-grained alignment techniques during pre-training requires a large amount of computation cost and time. 
Thus, a question naturally arises: \textit{Could we fine-tune the pre-trained models adaptive to downstream tasks with alignment objectives and achieve comparable or better performance?}
In this paper, we propose a new model architecture with alignment-enriched tuning (dubbed AETNet) upon pre-trained document image models, to adapt downstream tasks with the joint task-specific supervised and alignment-aware contrastive objective. 
Specifically, we introduce an extra visual transformer as the alignment-ware image encoder and an extra text transformer as the alignment-ware text encoder before multimodal fusion. 
We consider alignment in the following three aspects: 1) document-level alignment by leveraging the cross-modal and intra-modal contrastive loss; 2) global-local alignment for modeling localized and structural information in document images; and 3) local-level alignment for more accurate patch-level information.
Experiments on various downstream tasks show that AETNet can achieve state-of-the-art performance on various downstream tasks. Notably, AETNet consistently outperforms state-of-the-art pre-trained models, such as LayoutLMv3 with fine-tuning techniques, on three different downstream tasks. 
Code is available at~\url{https://github.com/MAEHCM/AET}.

\end{abstract}

\section{Introduction}
Self-pretraining techniques aiming to learn generic representations have recently proved to be highly effective for document image understanding. 
Notably, transfer learning based on pre-trained document image models yields strong performance on various document related downstream tasks~\cite{xu2020layoutlm,xu-etal-2021-layoutlmv2,garncarek2021lambert,hong2022bros,wu2021lampret,Li2021StructuralLMSP,li2021selfdoc,li2021structext,lee2022formnet, huang2022layoutlmv3, li2022dit}. 
A typical self-pretraining method in document image understanding is to pre-train a model on a large amount of pairs of the document images and OCR texts with layout information underlying the constraints of unsupervised losses. 
Downstream tasks then usually leverage the pre-trained weights for initialization. After that, the initialized model is fine-tuned for a downstream task using a task-specific supervised objective.  

\begin{figure}[t]
  \centering
  \label{fig:fig1}
  \includegraphics[width=0.7\linewidth]{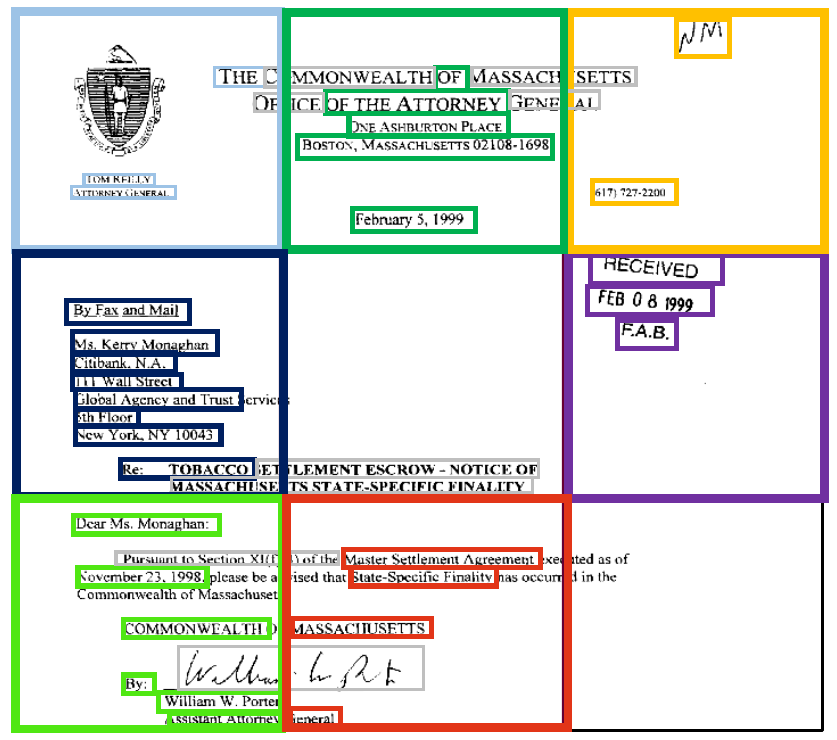}
  \caption{A document image example containing rich contextual text and layout information.}
  \label{fig:example}
\vspace{-10pt}
\end{figure}

As shown in Figure \ref{fig:example}, a document image contains rich contextual text and structural information, requiring fine-grained interaction modeling between image and text.
Although self-supervision achieves large progress in document image related-tasks, most existing pre-trained document image models are trained with coarse self-supervised losses, which ignore fine-grained interaction modeling between image and text. 
Inspired by this, DocFormer~\cite{Appalaraju_2021_ICCV} learns to reconstruct image pixels through a CNN decoder,
SelfDoc~\cite{li2021selfdoc} proposes to regress the masked region feature, and
the latest model LayoutLMv3~\cite{huang2022layoutlmv3} introduces a word-patch alignment objective by reconstructing masked patch tokens of the visual modality.
Although alignment between image and text has shown promising improvements on large-scale self-supervised pre-trained document image models, investigating more effective or finer-grained alignment techniques during the pre-training requires huge computational cost and time. 
Thus, a question naturally arises: \textit{Could we fine-tune the pre-trained models adaptive to downstream tasks with more alignment objectives and achieve comparable or better performance?}

\begin{figure*}[t]
  \centering
  \includegraphics[width=0.95\linewidth]{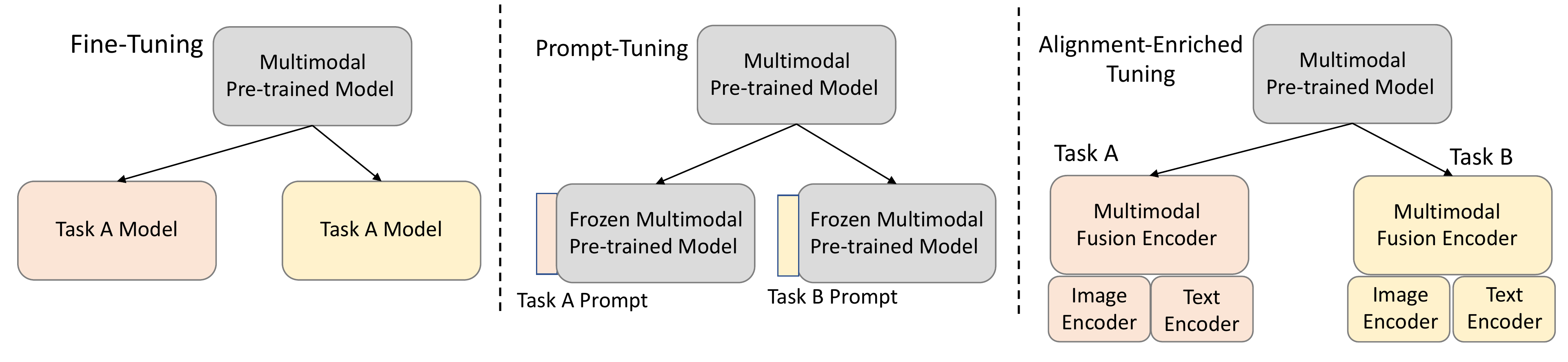}
  \caption{Fine-tuning requires tuning the entire pre-trained model for each downstream task. Prompt-tuning only requires tuning a trivial amount of extra parameters for each downstream task. Alignment-enriched tuning adds an extra alignment-aware image encoder and an extra alignment-aware text encoder before the multimodal pre-trained model.  Alignment-enriched tuning trains the entire parameters together for each downstream task.}
  \label{fig:tuning}
\vspace{-10pt}
\end{figure*}

In recent few years, with the wide success of large-scale pre-trained models, a range of tuning techniques has arisen to adapt these general-purpose models to downstream tasks~\cite{howard2018universal, jiang2019smart, houlsby2019parameter, gururangan2020don, xu2021raise, liu2021p, liu2021gpt, li2021prefix}.
The model tuning strategy (i.e., fine-tuning)~\cite{howard2018universal} tunes all model parameters during adaptation.
To improve training efficiency, adapters~\cite{houlsby2019parameter} and prompt tuning~\cite{li2021prefix, liu2021gpt, liu2021p} tune pre-trained models by a trivial amount of parameters, but they may suffer a small performance degradation compared with fine-tuning. 
In addition, task-adaptive pre-training~\cite{gururangan2020don} continues to train the pre-trained models with the unlabeled data of specific tasks. 
However, it is inconvenient to incorporate alignment loss into existing tuning strategies due to the limitation of fixed model architectures of pre-trained models. 

To enable the pre-trained models to have the ability of modeling alignment, ALBEF~\cite{li2021align} and TCL~\cite{yang2022vision} encode the document image and text independently with an extra image encoder and an extra text encoder before fusion by a multimodal encoder. 
The model is then pre-trained with alignment-aware strategies to improve the learned features' expressiveness, which is essential for joint multi-modal representation learning. 
However, as mentioned before, investigating more effective or finer-grained alignment techniques during the pre-training stage requires much computation cost and time.

To bridge the above research gap, in this paper, we propose a new model architecture with alignment-enriched tuning (termed AETNet), which tunes pre-trained document image models adaptively to enable downstream tasks with the joint task-specific supervised and alignment-aware contrastive objective. 
As shown in Figure~\ref{fig:tuning}, compared with fine-tuning and prompt-tuning, in our proposed AETNet, we introduce an extra visual transformer as the alignment-ware image encoder and an extra text transformer as the alignment-ware text encoder before multimodal fusion. 
We consider alignment in the following three aspects: 1) document-level alignment by leveraging the cross-modal and intra-modal contrastive loss; 2) global-local alignment for modeling localized and structural information in document images; and 3) local-level alignment for more accurate patch-level information.

We evaluate our AETNet method on various downstream document image understanding tasks, including FUNSD~\cite{jaume2019funsd} for form understanding, CORD~\cite{park2019cord} for receipt Understanding, DocVQA~\cite{mathew2021docvqa} for document visual question answering, and a sampled subset RVL-CDIP-1 from RVL-CDIP~\cite{harley2015icdar} for document image classification. 
In terms of performance, the proposed AETNet method  consistently outperforms existing state-of-the-art pre-trained models with fine-tuning on various downstream tasks.
Notably, with the help of alignment-enriched tuning, AETNet achieves better performance compared with general fine-tuning and prompt tuning strategies. 
We also carry out ablation studies with detailed analysis to investigate the effectiveness of each alignment loss in AETNet. Lastly, we conduct case studies over real examples from the FUNSD task to show that AETNet benefits from the alignment modeling. 


\section{Related Work}

\begin{figure*}[t]
  \centering
  \includegraphics[width=1.0\linewidth]{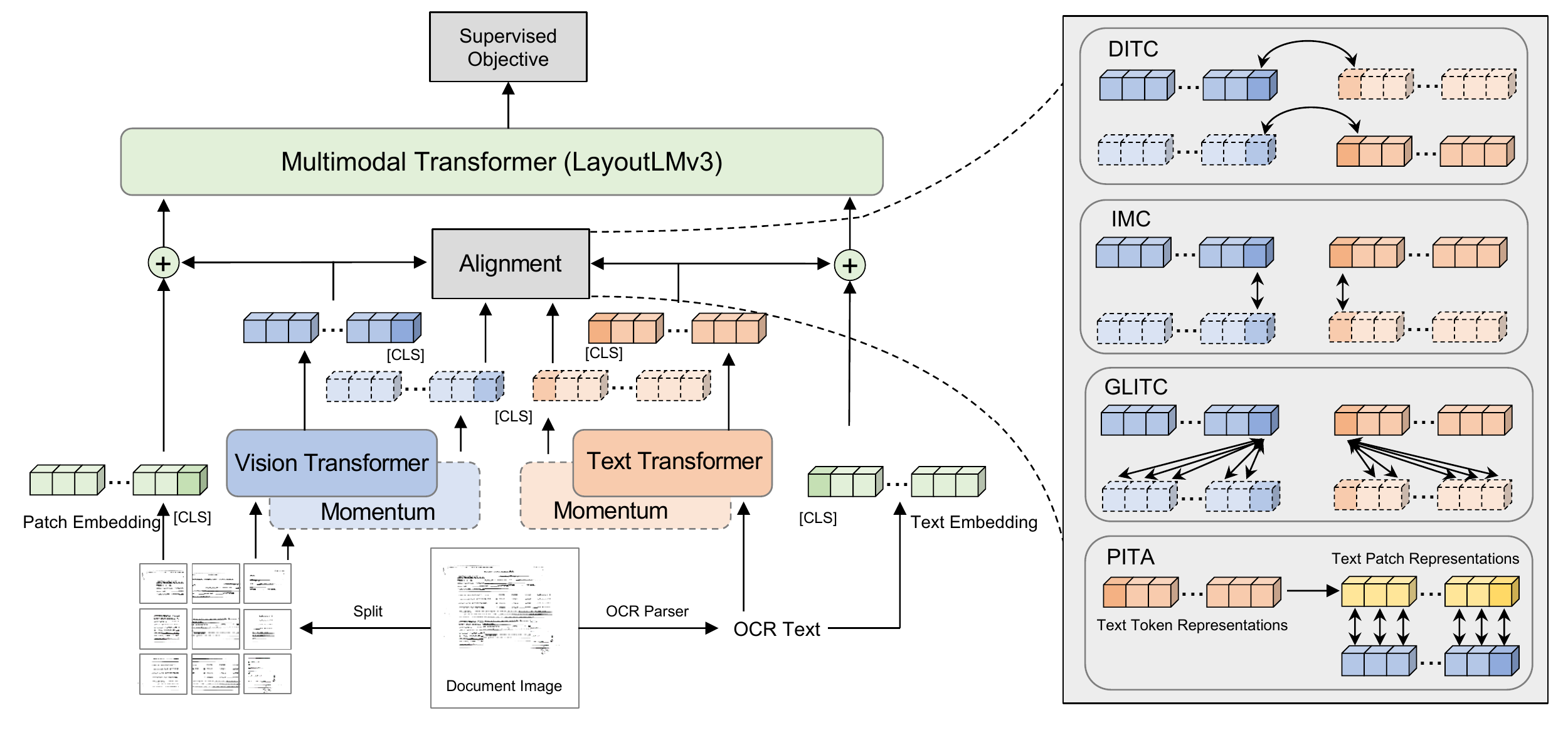}
  \caption{An overview of the framework of AETNet. 
  The model architecture consists of a vision transformer, a text transformer, and a fusion encoder. 
  Image and text encoders have their paired momentum encoder updated by the momentum-based moving average. 
  The alignment objective contains four contrastive losses (i.e., DITC, IMC, GLITC and PITA) for both alignment-enriched cross-modal and intra-modal representation learning.}
  \label{fig:overall}
\vspace{-10pt}
\end{figure*}

\subsubsection{Multimodal Pre-training}

Multimodal self-supervised pre-training has been successfully applied in document images through effectively leveraging image, layout, contextual text information~\cite{xu2020layoutlm,xu-etal-2021-layoutlmv2,garncarek2021lambert,Powalski2021GoingFB,wu2021lampret,Li2021StructuralLMSP,li2021selfdoc,Appalaraju_2021_ICCV,li2021structext,wang2022LiLT,gu2022xylayoutlm, kim2022donut}.
LayoutLM and its following works consider the layout information as a type of two-dimensional positional vectors and fuse their transformed vectors with text embeddings for the multimodal pre-trained model~\cite{xu2020layoutlm,Li2021StructuralLMSP,hong2022bros,lee2022formnet}. 
Some works extract CNN grid features~\cite{xu-etal-2021-layoutlmv2,Appalaraju_2021_ICCV} and some~\cite{xu2020layoutlm,Powalski2021GoingFB,li2021selfdoc,gu2021unidoc} rely on object detectors to extract region features.
However, these works are either limited by heavy computation bottleneck or require region supervision.
Recently researchers make many efforts to overcome the above limitations caused by CNN. Inspired by Vision Transformer (ViT)~\cite{dosovitskiy2020vit}, most rely on separate self-attention networks to learn visual features to reduce computational cost~\cite{xue2021probing,li2021align,dou2021empirical}.
For instance, ViLT~\cite{kim2021vilt}, one of works in vision-and-language pre-training utilizing ViT, learns visual features with a lightweight linear layer and significantly reduce the model parameters and running time.
Following ViLT, LayoutLMv3~\cite{huang2022layoutlmv3} is the first work to take advantage of patch-level embeddings without CNNs in Document images. For modeling fine-grained interaction and alignment between document image and their OCR text, we introduce alignment-aware ViT~\cite{dosovitskiy2020vit} and RoBERRTa~\cite{liu2019roberta} encoders before fusion.

\subsubsection{Tuning Techniques}

Fine-tuning is the useful paradigm for tuning
large pretrained language models for downstream tasks~\citep{devlin2019bert}.
In recent few years, different fine-tuning techniques have been proposed. Model tuning~\cite{howard2018universal}  requires tuning the entire pre-trained model for each downstream task. Mixout~\cite{lee2019mixout} randomly replaces part of
the model parameters with the pre-trained weights
during fine-tuning.
Child-Tuning~\cite{xu2021raise} updates parameters within the child network via a gradient mask.
There are also studies focusing on parameter-efficient fine-tuning, such as adapter-based methods~\cite{houlsby2019parameter}, and the prompt tuning methods~\cite{li2021prefix, liu2021gpt, liu2021p}.
These works tune pre-trained models by a trivial amount of parameters while they may suffer a small performance degradation compared with fine-tuning. Beyond, it is inconvenient to incorporate alignment loss into these existing tuning strategies due to the limitation of fixed model architectures of pre-trained models.

\section{Methodology}

In this section, we present a new model architecture with the proposed alignment-enriched tuning (AETNet) for the transfer of the patch-level pre-trained models in document images to downstream tasks. We first introduce the model architecture, followed by the  tuning process. Then we detail the tuning objectives including downstream task specific supervised objective, document-level image-text contrastive objective, intra-modal contrastive objective, global-local image-text contrastive objective, and patch-level image-text alignment objective.


\subsection{Model Architecture}
The model architecture of our method AETNet is shown in Figure~\ref{fig:overall}. 
The model architecture consists of a large-scale patch-level multimodal pre-trained model, an extra alignment-aware image encoder, and an extra alignment-aware text encoder.
For the multimodal pre-trained model, we employ LayoutLMv3~\cite{huang2022layoutlmv3} as our pre-trained model because of not only its state-of-the-art performance on downstream tasks but also its patch-level image encoder that is suitable for the AETNet process. 
We use a RoBERTa~\cite{liu2019roberta} as the extra alignment-aware text encoder. Likewise, we use a ViT~\cite{dosovitskiy2020vit} as the extra alignment-aware image encoder, which is initialized using weights pre-trained on ImageNet-1k from DeiT~\cite{touvron2020deit}. The implementation details of extra text and image encoders, are described in Experiment Section. 

\subsection{Tuning Process}

In the downstream tuning process, our model is trained with the joint loss of downstream task specific objective and the proposed alignment-enriched objective.
As shown in Figure~\ref{fig:overall}, an input document image $I$ is first encoded into a sequence of patch-level visual representation vectors: $\{v_{cls}, v_1, ..., v_N\}$ through the alignment-aware ViT, where $cls$ denotes the special token [CLS] and $N$ is the number of patches.
Meanwhile, the input text (i.e., the textual content of this document image, which is obtained by the open-source OCR toolkit, Tesseract\footnote{\url{https://github.com/tesseract-ocr/tesseract}}) is fed into the alignment-aware RoBERTa to be transformed into a sequence of text token representation vectors: $\{w_{cls}, w_1, ..., w_L\}$, where $L$ is the length of the input text tokens. 
Before fusion with LayoutLMv3, we compute alignment-aware losses based on the obtained representations in terms of alignment-aware objectives to let the obtained representations preserve alignment information. 
Then, we derive the patch-level input image embeddings for LayoutLMv3 by the sum of obtained alignment-enriched image representation vectors and original input image embeddings of LayoutLMv3.
Likewise, input text embeddings for LayoutLMv3 are derived by the sum of obtained alignment-enriched text representation vectors and original input text embeddings of LayoutLMv3.
Lastly, LayoutLMv3 transforms the newly fused embeddings into output hidden representation for computing supervised task-specific loss. 

\subsection{Tuning Objectives}

The full tuning objective of AETNet consists of five objectives: the downstream task-specific subjective objective (SO), global-level cross-modal alignment (GCMA), global-level intra-modal contrastive (GIMC), Global-Local MI Maximization (GLMI), and Local-level Cross-Modal Alignment (LLCMA), shown as below:
\begin{equation}
    \mathcal{L}_{aet}= \mathcal{L}_{so}+\mathcal{L}_{ditc}+\mathcal{L}_{imc}+\mathcal{L}_{glitc}+\mathcal{L}_{pita}.
\end{equation}
In the following, we elaborate each objective in details.

\subsubsection{Supervised Objective (SO)}

In the vanilla fine-tuning framework, the model is first initialized with the parameters of large pre-trained models (LPM). Then, the initialized model is fine-tuned for a certain downstream task using the task-specific objective on the corresponding dataset.
Unlike vanilla fine-tuning, our model consists of three modules: LPM, the alignment-aware image encoder, and the alignment-aware text encoder. 
We train these three modules together in our model with the task-specific objective: $\mathcal{L}_{so} (I, T, Y)$, where image $I$ and OCR text $T$ are the input and $Y$ is for the ground-truth labels.
We take the semantic entity labeling task~\cite{jaume2019funsd, park2019cord}, one of the downstream tasks mentioned in Experiment Section, as an example. $L_{so}$ is the cross entropy loss based on OCR tokens' predictions and their corresponding ground-truth labels.

\subsubsection{Document-Level Image-Text Contrastive (DITC)} The purpose of DITC is to learn better document(global)-level alignment cross-modality through a contrastive loss, which pulls representations of matched image-text pairs to be close and pushes the unmatched ones apart. 
To model global information, we apply similarity functions $sim(v_{cls},\tilde{w}_{cls})=f_{img}(v_{cls})^\top f_{txt}(\tilde{w}_{cls})$ and $sim(w_{cls},\tilde{v}_{cls})=f_{text}(w_{cls})^\top f_{img}(\tilde{v}_{cls})$ for the DITC loss, where $f_{img}$ and $f_{txt}$ are two projection heads for mapping hidden vectors into low-dimensional representations in the contrasitve loss space and $v_{cls}$ and $w_{cls}$ are two [CLS] vectors from alignment-aware image and text encoders, respectively. 
Next, we refer to MoCo~\cite{he2020momentum} and ALBEF~\cite{li2021align} to maintain two memory queues (i.e., $\tilde{I}_{cls}=\{\tilde{v}^1_{cls}, ..., \tilde{v}^K_{cls}\}$ for images, $\tilde{T}_{cls}=\{\tilde{w}^1_{cls}, ..., \tilde{w}^K_{cls}\}$ for text) to record the most recent K image and text representations from the momentum alignment-aware encoders. 
The document-level contrastive loss for a pair of document image $I$ and OCR text $T$ can be defined as follows:
\begin{equation}
\begin{aligned}
    \mathcal{L}_{cl}(v_{cls}, \tilde{w}_{cls}, \tilde{T}_{cls}) = -\log\frac{\exp (sim(v_{cls},\tilde{w}_{cls}) / \tau)}{\sum_{k=1}^K \exp (sim(v_{cls},\tilde{w}^{k}_{cls})/ \tau)},
    \end{aligned}
\end{equation}
where $\tau$ is the learnable  temperature rate. Considering image-to-text and text-to-image together, DITC loss is defined as:
\begin{equation}
\begin{aligned}
    \mathcal{L}_{ditc} = \frac{1}{2}  \mathbb{E}_{(I,T)\sim B}\big[\mathcal{L}_{cl}(v_{cls}, \tilde{w}_{cls}, \tilde{T}_{cls}) + \\ \mathcal{L}_{cl}(w_{cls},\tilde{v}_{cls}, \tilde{I}_{cls})\big],
    \end{aligned}
\end{equation}
where $B$ is a batch of image-text pairs. In addition, We follow ALBEF~\cite{li2021align} to guide the training of DITC loss by pseudo-targets generated by the momentum model.

\subsubsection{Intra-Modal Contrastive (IMC)}

The goal of IMC is to learn more accurate representations within the same modality. 
For the visual modality, we generate positive latent image representation $\{v^+_{cls}, v^+_1, ..., v^+_N\}$ for the anchor augmented image $I^+$ by feeding the anchor image into the momentum alignment-aware image encoder. For the text modality, the positive latent representation $\{w^+_{cls}, w^+_1, ..., w^+_L\}$ for the anchor text $T$ is produced by the momentum model in the same way as image samples. Two momentum queues used for the DITC loss also provide negative samples for the IMC loss. The IMC loss is define as follows:
\begin{equation}
\begin{aligned}
    \mathcal{L}_{imc} = \frac{1}{2} \mathbb{E}_{(I,T)\sim B}\big[\mathcal{L}_{cl}(v_{cls}, v^+_{cls}, \tilde{I}_{cls}) +\\ \mathcal{L}_{cl}(w_{cls}, w^+_{cls}, \tilde{T}_{cls})\big].
\end{aligned}
\end{equation}

\subsubsection{Global-Local Image-Text Contrastive (GLITC)}

Inspired by TCL~\cite{yang2022vision}, we introduce the GLITC loss to capture localized and structural information within the single modality by modeling interactions between the document-level representation and local regions. 
Specifically, for the visual modality, we use global representation $v_{cls}$ and compute the contrastive loss with momentum image patch representations$\{{\widetilde{v}}_{2},\ldots,{\widetilde{v}}_{N}\}$. Likewise, for the the text modality, we use the text [CLS] representation $w_{cls}$ and momentum text token representations $\{{\widetilde{w}}_{2},\ldots,{\widetilde{w}}_{L}\}$ for computing loss. We encourage to model interactions between global information and lcoal information within the same modality by minimizing the following two-side interaction contrastive loss:
\begin{equation}
\begin{aligned}
   \mathcal{L}_{glitc} = \frac{1}{2}\mathbb{E}_{(I,T)\sim B}\big[\frac{1}{N}\sum^{N}_{i=1}\mathcal{L}_{cl}(v_{cls}, \tilde{v}_i, I^-_B) + \\ \frac{1}{L}\sum^{L}_{j=1}\mathcal{L}_{cl}(w_{cls}, \tilde{w}_j, T^-_B)\big],
\end{aligned}
\end{equation}
where $I^-_B$ and $T^-_B$ are negative image patch and text token representation vectors, respectively. These negatives are from other data examples in the same batch $B$.


\subsubsection{Patch-Level Image-Text Alignment (PITA)}
Although the introduced DPITC loss has the ability to capture certain cross-modal localized and structural information in the input, it ignores patch-level alignment between patch-level images and patch-level contextual text. 
Patch-level alignment is critical to learning more accurate and finer-grained representations for more accurate document understanding. 
Therefore, we propose a novel patch-level alignment loss as a complement to the previous three contrastive losses. 
As shown in Figure~\ref{fig:example}, to compute the patch-level alignment loss, we first find text tokens matching image patches by a rule-based strategy that locates text in the image patches by layout information obtained from the OCR toolkit. 
Then, we average the representation vectors of the matched text tokens to derive patch-level text representations for the matched text $\{t_1, t_2, ..., t_N\}$.  
The PITA loss is defined below:
\begin{equation}
   \mathcal{L}_{pita} = -\frac{1}{N}\big( \sum^{N}_{i=1}
   \frac{\langle v_i,t_i \rangle}{||v_i||\cdot||t_i||} \big).
\end{equation}

\section{Experiment}
\label{sec:exp}
\subsection{Experiment Setting}
\subsubsection{Model Configures}

Since we aim to investigate the impact of the proposed fine-grained alignment losses during the tuning process, we thus adopt the patch-level multimodal pre-trained model (LayoutLMv3~\cite{huang2022layoutlmv3}) instead of document-level pretrained models as our fusion model.
LayoutLMv3 is pre-trained on 11 million document images of a large IIT-CDIP~\cite{harley2015icdar} dataset. 
LayoutLMv3$_{\text{BASE}}$ uses a 12-layer Transformer~\cite{vaswani2017attention} Encoder with 12 self attention heads, 768 hidden size, and 3,072 intermediate size for feed-forward network.
LayoutLMv3$_{\text{BASE}}$ uses a 24-layer Transformer~\cite{vaswani2017attention} Encoder with 16 self attention heads, 1,024 hidden size, and 4,096 intermediate size for feed-forward network. Patch size in LayoutLMv3 is 16. The alignment-aware ViT and RoBERTa encoders before fusion are initialized with the model parameters released by DeiT~\cite{touvron2020deit} and RoBERTa~\cite{liu2019roberta}, respectively. The size of the momentum queues of all encoders is 65536.

\subsubsection{Downstream Tasks}
We evaluate the proposed alignment-enriched tuning (AET) method on three document multimodal downstream tasks as follows:

\noindent\textbf{Form and Receipt Understanding} is a sequential labeling task, which aims to assign a label to each word. 
Form and receipt understanding requires an understanding of visual information and textual content extracted from structural forms in document images. 
We follow fine-tuning settings in LayoutLMv3 to evaluate AET on two public datasets, i.e., FUNSD~\cite{jaume2019funsd} and CORD~\cite{park2019cord}. 
FUNSD is a dataset sampled from the RVL-CDIP dataset~\cite{harley2015icdar} about noisy scanned form understanding. 
It consists of 199 documents (149 documents for training and 50 documents for testing) and 9,707 semantic entities.
CORD is a receipt key information extraction dataset, including 1,000 receipts and 30 semantic labels defined under 4 categories, where 800 samples are used for training, 100 for validation, and 100 for testing.

\noindent\textbf{Document Visual Question Answering} is to predict the answer given an document image and a question, which requires models understanding the knowledge in documents and learning to  reason over documents to answer. We follow the task formulation in LayoutLMv3, considering this task as an extractive QA problem. 
Specifically, the model is required predicting start and end positions of the answer in the document. This is a binary classification over each text token. 
We follow the official partition of the DocVQA~\cite{mathew2021docvqa} dataset, which consists of 10,194/1,286/1,287 images with 39,463/5,349/5,188 questions for training/validation/test, respectively.

\noindent\textbf{Document Image Classification} is a document classification task aiming to predict the category of a given document, which requires of an understanding of document contents. Due to the limitations of our servers, we sampled a subset from RVL-CDIP~\cite{harley2015icdar}, termed RVL-CDIP-1 , including 10,000 data examples with 16 categories from the RVL-CDIP datasets of 400,000 document images as our evaluation dataset. RVL-CDIP-1 is divided into 8000 training samples, 1000 validation samples, and 1000 test samples.

\subsection{AETNet on Downstream Tasks}

\begin{table}[t]
    \centering
    \small
    
    \resizebox{0.50\textwidth}{!}{
    \begin{tabular}{l|cc}
    \toprule
      \multirow{2}{*}{\bf Model} & \bf FUNSD & \bf CORD     \\
       & \bf F1$\uparrow$  & \bf F1$\uparrow$  \\
     \midrule
      $\textrm{BERT}_{base} $~\cite{devlin2019bert}    & 60.26 & 89.68 \\
      $\textrm{RoBERTa}_{base} $~\cite{liu2019roberta}  & 66.48 & 93.54 \\
      $\textrm{BROS}_{base} $~\cite{hong2022bros}   & 83.05 & 95.73 \\
      $\textrm{UDoc}$~\cite{gu2021unidoc}  & 87.93 & 98.94$^\dagger$ \\
      $\textrm{LayoutLMv2}_{base} $~\cite{xu-etal-2021-layoutlmv2}    & 82.76 & 94.95 \\
      $\textrm{DocFormer}_{base} $~\cite{Appalaraju_2021_ICCV}   & 83.34 & 96.33  \\
    LayoutLMv3$_{base}$~\cite{huang2022layoutlmv3}    & 89.82 & 95.97 \\
     AETNet$_{base}$ (Ours)   & \textbf{91.55} & \textbf{97.04}\\
     
     \midrule
     
      $\textrm{BERT}_{large}  $~\cite{devlin2019bert}   & 65.63 & 90.25 \\
      $\textrm{RoBERTa}_{large} $~\cite{liu2019roberta}   & 70.72 & 93.80\\
     $\textrm{BROS}_{large}  $~\cite{hong2022bros} & 84.52 & 97.40 \\
      $\textrm{LayoutLMv2}_{large} $~\cite{xu-etal-2021-layoutlmv2}  & 84.20 & 96.01  \\
      $\textrm{DocFormer}_{large}  $~\cite{Appalaraju_2021_ICCV}  &  84.55 & 96.99  \\
     LayoutLMv3$_{large}$~\cite{huang2022layoutlmv3}   & 90.94 & 97.01  \\
     AETNet$_{large}$ (Ours) &\textbf{92.33} & \textbf{97.52} \\
     \bottomrule
    
    \end{tabular}
    }
    
    \caption{
    Performance comparison with  existing published pre-trained models with fine-tuning on FUNSD and CORD datasets. The results of LayoutLMv3 are based on our implementations with the released
    UniLM~\cite{dong2019unified}. The score$^\dagger$ is not directly comparable to other scores.
    }
    \label{tab:seqlabeling}
    \vspace{-10pt}
\end{table}

The detailed description of hyper-parameters, including running epochs, learning rate, batch size, and optimizer, for our method on three downstream tasks and four datasets, are referred to in the supplementary.

\subsubsection{Evaluation on Form and Receipt Understanding}

We compare the proposed model architecture with alignment-enriched tuning (AETNet) to three types of self-supervised pre-training approaches using fine-tuning: (1) Pre-trained models only using text modality. 
BERT~\cite{devlin2019bert} and RoBERTa~\cite{liu2019roberta} is based on Transformer architecture and only use text information; (2) Pre-trained models with text and layout modalities. BROS~\cite{hong2022bros} incorporates layout information by encoding relative layout positions; (3) Pre-trained models with text, layout and image modalities. UDoc~\cite{gu2021unidoc} adopts object proposals from document images and concatenate region features and text embeddings. LayoutLMv2~\cite{xu-etal-2021-layoutlmv2} use a CNN network to extract image features and then feed it with layout and text information into multimodal Transformer. DocFormer~\cite{Appalaraju_2021_ICCV} extract image features with CNN. LayoutLMv3~\cite{huang2022layoutlmv3} replaces CNN backbones with patch-level linear embedding layers.

Table~\ref{tab:seqlabeling} reports the results of the comparison AETNet to fine-tuning for existing published pre-trained models on FUNSD and CORD datasets.  
Our APTNet achieves state-of-the-art performance while outperforming the previous fine-tuning based methods. 
For transferring the pre-trained model to the downstream tasks, we compared AETNet$_{base}$ to the reported fine-tuning for LayoutLMv3$_{base}$~\cite{huang2022layoutlmv3} and our implemented fine-tuning for LayoutLMv3$_{base}$. AETNet$_{base}$ improves $+1.25$ and $+1.73$ on FUNSD and $+0.51$ and $+1.07$ on CORD, revealing the necessity of conducting document-level and patch-level intra-modal and cross-modal alignment after pre-training.

\subsubsection{Evaluation on Document Visual Question Answering}
\begin{table}[t]
    \centering
    \small
    \begin{tabular}{l|c}
    \toprule
     \multirow{2}{*}{\bf Model} & \bf DocVQA    \\
     & \bf ANLS$\uparrow$ \\
     \midrule
      $\textrm{BERT}_{base} $~\cite{devlin2019bert}  & 63.72  \\
      $\textrm{RoBERTa}_{base} $~\cite{liu2019roberta} &  66.42  \\
      $\textrm{LayoutLMv2}_{base}$~\cite{xu-etal-2021-layoutlmv2}    &  78.08 \\

      LayoutLMv3$_{base}$~\cite{huang2022layoutlmv3}    & 78.76 \\
      AETNet$_{base}$ (Ours)  & \bf 79.73 \\
     

     
     \bottomrule
    
    \end{tabular}
    \caption{
    \small
    Performance comparison on DocVQA dataset.
    }
    \label{tab:docvqa}
    \vspace{-10pt}
\end{table}

\begin{figure}[t]
  \centering
  \includegraphics[width=1.0\linewidth]{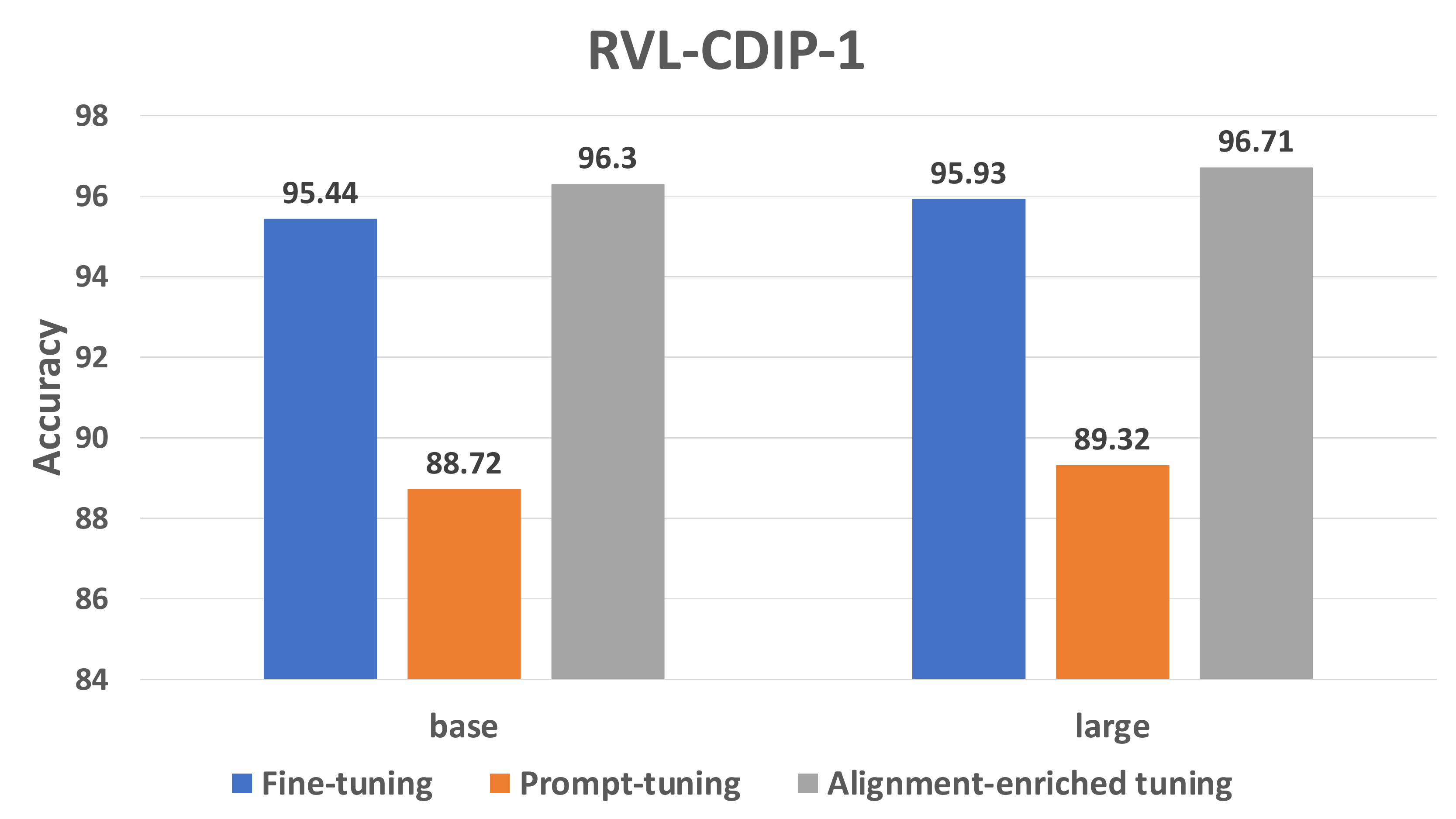}
  \caption{Performance comparison on the sampled subset RVL-CDIP-1 dataset.}
  \label{fig:cls}
\vspace{-10pt}
\end{figure}

\begin{table}[t]
    \small
    \resizebox{0.48\textwidth}{!}{
    \aboverulesep = 0.55mm
    \belowrulesep = 0.55mm
	\centering	
	\begin{tabular}	{l  l |  c  c  }
		\toprule	 	
	 \bf \#Pre-train & \multirow{2}{*}{\bf Training tasks} & \bf FUNSD &  \bf CORD\\
	 \bf Models &  & \bf F1$\uparrow$  & \bf F1$\uparrow$  \\
	 	 \midrule
	 \multirow{4}{*}{LayoutLMv3$_{base}$}	 & Fine Tuning &89.82 &95.97 \\
	 	 & P-Tuning~\cite{liu2021p} &83.75 &87.37 \\ 
	 	 & AETNet  &\textbf{91.55} &\textbf{97.04} \\ 	 
	
		\bottomrule
	\end{tabular}
}
	\caption
	{
	\small	
Comparison to different adaptation approaches on FUNSD and CORD datasets. 
Fine Tuning: All pre-trained parameters are tuned with the supervised objective. 
P-Tuning: Prompt parameters of each layer are tuned with the supervised objective. 
AETNet: Pre-trained parameters and extra alignment-aware image and text encoders are tuned with the supervised objective and our proposed alignment loss. All results are based on our implementation.
	}
	\label{tab:ablation_tuning}
\vspace{-10pt}
\end{table}

\begin{table}[t]
    \centering
    \small
    
    \resizebox{0.47\textwidth}{!}{
    \begin{tabular}{lcc}
    \toprule
    
     \multirow{2}{*}{\bf Tuning Objective} & \bf FUNSD & \bf CORD  \\
     & \bf F1$\uparrow$ & \bf F1$\uparrow$ \\
     
     \midrule
     
     LayoutLMv3$_{base}$ (Fine-tuning) &89.82 &95.97 \\
    \midrule
    AETNet$_{base}$ (w/ Supervised Objective) &89.77  &96.30   \\
    \quad + DITC &90.44  &96.48\\
    \quad + IMC &90.45  &96.61   \\
    \quad + GLITC &90.18  &96.37    \\
    \quad + PITA &90.69  &96.78   \\
    AETNet$_{base}$ & \textbf{91.55} &\textbf{97.04} \\
    \bottomrule
    \end{tabular}
    }
    \caption{
    \small
    Ablation study of each loss in the AETNet on FUNSD and CORD datasets. The F1 is reported. DITC, IMC, DLITC, and PITA means Document-Level Image-Text Contrastive, Intra-Modal Contrastive, Global-Local Image-Text, and Patch-Level Image-Text Alignment, respectively. All results are obtained by our implementation.
    }
    \label{tab:ablation_obj}
    \vspace{-10pt}
\end{table}

\begin{figure*}[t]
  \centering
  \includegraphics[width=0.89\linewidth]{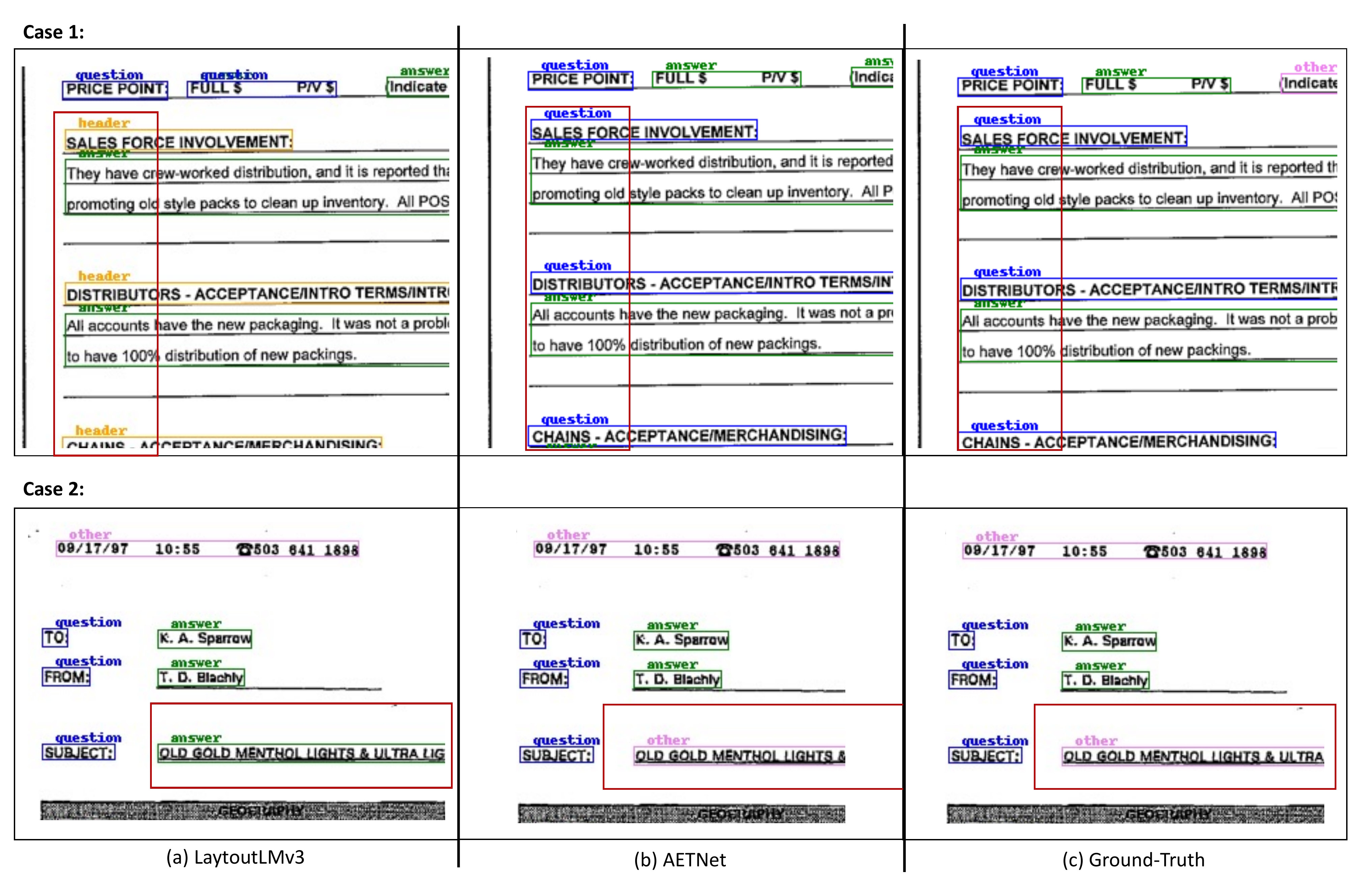}
  \caption{Visualization of two cases  on FUNSD  , which are predicted by LayoutLMv3 (Column (a)) and AETNet (Column (b). Column (c) is the  Ground-truth annotations.
}
  \label{fig:case}
\vspace{-10pt}
\end{figure*}

Table~\ref{tab:docvqa} reports the performance comparison on the DocVQA dataset. We compare AETNet$_{base}$ to methods only requiring text modality, i.e., BERT$_{base}$, RoBERTa$_{base}$, and methods requiring image and text as inputs, i.e., LayoutLMv2$_{base}$~\cite{xu-etal-2021-layoutlmv2} and LayoutLMv3$_{base}$~\cite{huang2022layoutlmv3}. AETNet$_{base}$ substantially outperforms existing baseline methods, achieving an absolute accuracy of $+0.97$ boost compared to LayoutLMv3$_{base}$.

\subsubsection{Evaluation on Document Image Classification}

Figure~\ref{fig:cls} shows the performance comparison on the sampled subset classification dataset, RVL-CDIP-1. We compare AETNet with the general fine-tuning and prompt-tuning strategies for LayoutLMv3 in both base and large model sizes. Although prompt-tuning is more parameter efficient, it cannot outperform fine-tuning and alignment-enriched tuning. Likewise, fine-tuning outperforms prompt tuning but is worse than alignment-enriched tuning. Overall, in terms of performance, AETNet can achieve desirable results on document image classification with the help of alignment modeling.

\subsection{Intrinsic Analysis}
To learn the effectiveness of each component in AETNet,
we first compare AETNet with the general fine-tuning strategy and a commonly-used prompt tuning strategy.
Then, we perform ablation study of each loss in the full objective on FUNSD and CORD datasets. 
Lastly, we do a case study to demonstrate the ability of AETNet to model alignment information.

\subsubsection{AETNet vs Other Tuning Methods}

As shown in Table~\ref{tab:ablation_tuning}, we compare AETNet with fine-tuning based LayoutLMv3$_{base}$ and P-tuning based LayoutLMv3$_{base}$ on FUNSD and CORD datasets. 
With the help of alignment-enriched tuning, AETNet achieves the best performance among these three tuning strategies. Specifically, AETNet achieves absolute improvements of $+1.73$ points over fine-tuning and $+7.79$ over prompt-tuning on FUNSD, and $+1.07$ points over fine-tuning and $+9.67$ over prompt-tuning on CORD. It indicates that alignment-enriched tuning has the ability to improve the quality of representations when transferring pre-trained models to downstream tasks.

\subsubsection{Ablation Study of each Loss in AETNet}

To learn the effectiveness of the newly proposed losses
(i.e., DITC, IMC, GLITC, and PITA) in improving the document multi-modal representation learning, we perform an ablation study of each each loss in AETNet on FUNSD and CORD datasets. 
Table~\ref{tab:ablation_obj} report the results. 
Firstly, it is worth mentioning that AETNet$_{base}$ using supervised objective is slightly worse than LayoutLMv3$_{base}$ using supervised objective.
In other words, adding more parameters during the fine-tuning process in this study cannot improve performance. This observation further verifies that downstream tasks benefit from alignment-enriched representations learned by our proposed alignment-enriched objective during the tuning process. 
Secondly, combining with any newly proposed loss in AETNet$_{base}$ improves the performance on FUNSD and CORD, which can be attributed to the consideration of document-level or patch-level alignment-enriched representations.



\subsubsection{Case Study}

Figure~\ref{fig:case} shows the visualization of two cases predicted by LayoutLMv3 and AETNet. 
In case 1, AETNet predict three entities within the reb box as ``question'' but LayoutLMv3 predict them as ``header''. The possible reason is that LayoutLMv3 makes the prediction based on the layout information rather than content text, which requires models to have more accurate interaction and alignment modeling ability to avoid this error. 
In case 2, LayoutLMv3 learns from two neighbors above it to predict the entity within the red box as ``answer'', which indicates that the model utilizes too much layout information to make predictions rather than based on understanding it.

\section{Conclusion}

This paper proposes a new model architecture with alignment-enriched tuning method named AETNet for transferring multimodal pre-trained models in document images. 
AETNet consists of an extra visual transformer as the alignment-ware image encoder and an extra text transformer as the alignment-ware text encoder, and a multimodal fusion encoder initialized by pre-trained model parameters. Based on this model architecture, AETNet tunes pre-trained models for downstream tasks with joint task-specific supervised and alignment-aware contrastive objective.
AETNet achieves consistent improvements over existing state-of-the-art pre-trained models with fine-tuning and different commonly-used tuning techniques over various downstream tasks, which indicates the effectiveness of our proposed AETNet.

\section*{Acknowledgments}
This work was supported in part by National Natural Science Foundation of China under Grants (No. 61976049).

\bibliography{aaai23}

\newpage

\appendix



\section{Supplementary Material}
Due to the limitation of writing space, we provide more analysis of our proposed AETNet in this supplementary material. More specifically, we include the following three folds:
\begin{itemize}
	\item 
	Detailed implementation of the proposed AETNet method on various downstream tasks.
	\item
	Additional intrinsic analysis of the impact of the number of patches after the average pooling and the impact of number of layers of the proposed alignment-ware image and text encoders.
	\item  
	More visualization of cases to demonstate the strong alignment ability of our proposed AETNet.
\end{itemize}

\subsection{Implementation Details}

This section describes the detailed implementation of the AETNet method on various downstream tasks.

For all downstream tasks, we run experiments using $2$ NVIDIA Tesla 3090 GPUs. For initialization of model parameters of AETNet$_{base}$ and AETNet$_{large}$, we use DeiT~\cite{touvron2020deit} to initialize the alignment-aware image encoder, RoBERTa~\cite{liu2019roberta} to initialize the alignment-aware text encoder, and LayoutLMv3~\cite{huang2022layoutlmv3} to initialize the multimodal fusion encoder. 

For the base model, we set the batch size and epochs for alignment-enriched tuning to be $4$ and $30$ respectively for all downstream tasks. We set the number of layers for the alignment-aware image encoder, the number of layers for the alignment-aware text encoder, and hidden size to be 12, 12, and 768, respectively. We follow LayoutLMv3$_{base}$ to set number of layers for the multimodal fusion encoder. 

For the large model, we set the batch size and epochs for alignment-enriched tuning to be $2$ and $60$ respectively for all downstream tasks. We set the number of layers for the alignment-aware image encoder, the number of layers for the alignment-aware text encoder, and hidden size to be 6, 6, and 1024, respectively. We follow LayoutLMv3$_{large}$ to set number of layers for the multimodal fusion encoder.

\subsubsection{Implementation Details on FUNSD}
We evaluate the proposed AETNet method on the FUNSD dataset for the form understanding task. We follow the official split of the FUNSD dataset~\cite{jaume2019funsd}, which consists of 199 documents (149 documents for training and 50 documents for testing) and 9,707 semantic entities. We adopt the AdamW optimizer. The initial learning rate is 3e-2 and weight decay here is 0.02. We use the learning rate scheduler, i.e., CosineLRScheduler, with the minimum learning rate of 2e-2. Warmup learning rate and epochs are 2e-5 and 20, respectively.

\subsubsection{Implementation Details on CORD}
For the receipt understanding task, we evaluate on the CORD dataset~\cite{park2019cord}, including 1,000 receipts and 30 semantic labels defined under 4 categories, where 800 samples are used for training, 100 for validation, and 100 for testing. We use AdamW with an initial learning rate of 5e-2, CosineLRScheduler with the minimum learning rate of 3e-2, and weight decay of 1. The warmup learning rate we use is 2e-5, and warmup epochs are 20. We tune models for 40 epochs.

\subsubsection{Implementation Details on DocVQA}

For the document visual question answering task, we conduct experiments on the DocVQA dataset~\cite{mathew2021docvqa}. It consists of 10,194/1,286/1,287 images with 39,463/5,349/5,188 questions for training/validation/test, respectively. A data example here includes a document image, a question, and an answer. We use open-source OCR Tesseract\footnote{\url{https://github.com/tesseract-ocr/tesseract}}) to obtain the bounding boxes for each document image. For efficiency, we employ fp16 rather than fp32. We train the model with the training batch size of 8, epochs of 10, learning rate of 3e-5, and same learning scheduler as our experiments on FUNSD with the minimum learning rate of 2e-5. We follow LayoutLMv3 For the rest of hyper-parameters.

\subsubsection{Implementation Details on RVL-CDIP-1}

For the document image classification task, we evaluate all experiments on a small-scale dataset termed RVL-CDIP-1, which is a sampled subset of the RVL-CDIP dataset~\cite{harley2015icdar}. We sample examples from the original datasetset according to the distribution of 16 categories. In the end, RVL-CDIP-1 contains 8000/1000/1000 data samples for the training/validation/test. We train the model with the training batch size of 4, epochs of 80, learning rate of 3e-2, and same learning scheduler as our experiments on FUNSD with the minimum learning rate of 2e-2.

\begin{figure}[t]
  \centering
  \includegraphics[width=1.0\linewidth]{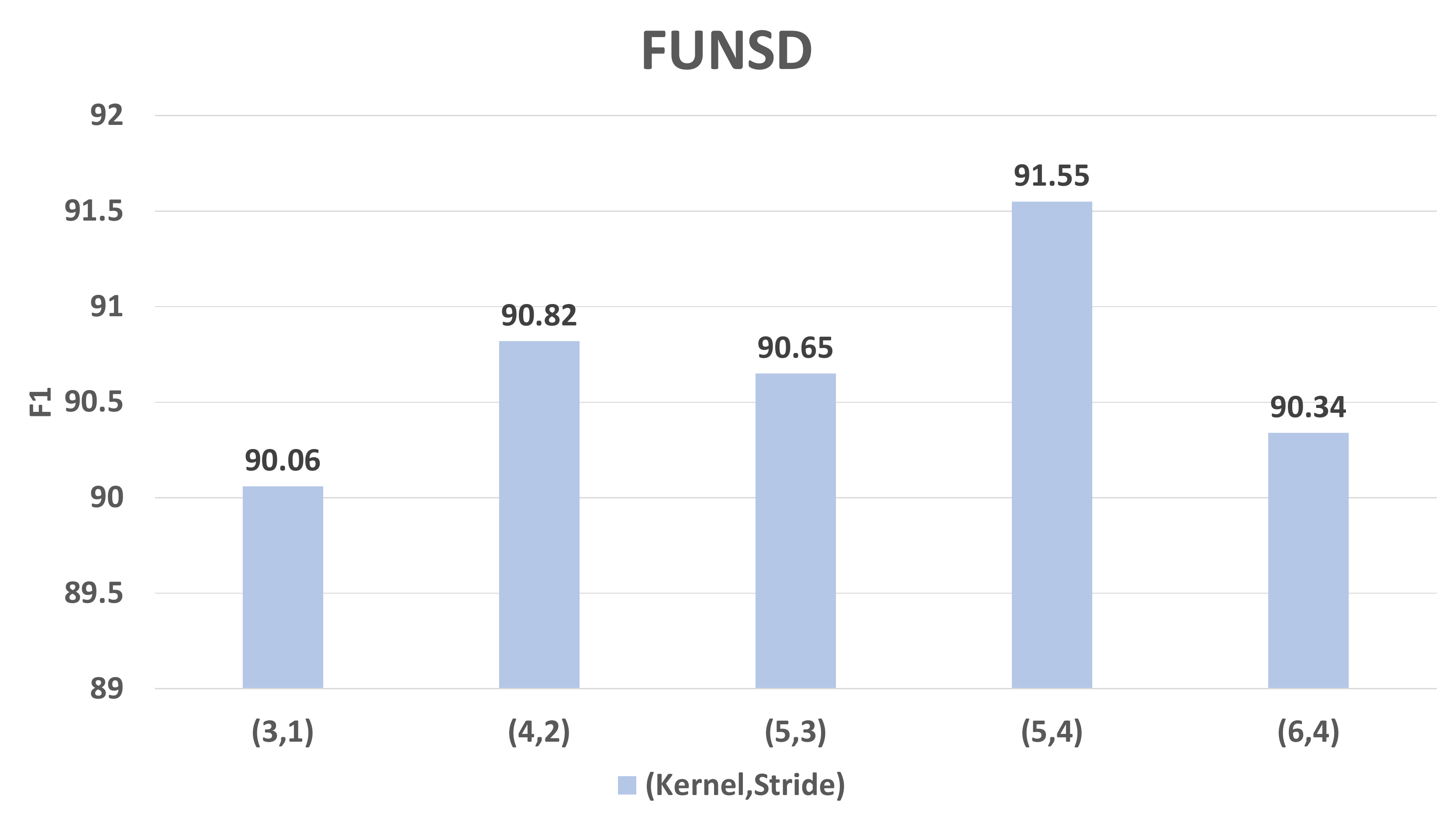}
  \caption{
  Performance (F1) comparison w.r.t. different kernel sizes and stride values for the average pooling over patch embeddings.
  }
  \label{fig:patches}
\end{figure}

\subsection{Additional Intrinsic Analysis}

\subsubsection{Impact of the Number of Patches}

In our all experiments, each document image is first split into 196 patches with a size of $14\times14$. 
Following TCL~\cite{yang2022vision}, we employ the global average pooling over the last layer of image and text patch embeddings for ruling out the probability that small document image patches may not contain enough text information. Figure~\ref{fig:patches} shows that AETNet benefits the most from $3\times3$ averaged patches with kernel size of 5 and stride value of 4, which indicates that there exist small document image patches which may not include text information or only include trivial and useless text information.

\subsubsection{Impact of Number of Layers of the Alignment-Aware Image Encoder}

\begin{figure}[t]
  \centering
  
  \includegraphics[width=0.86\linewidth]{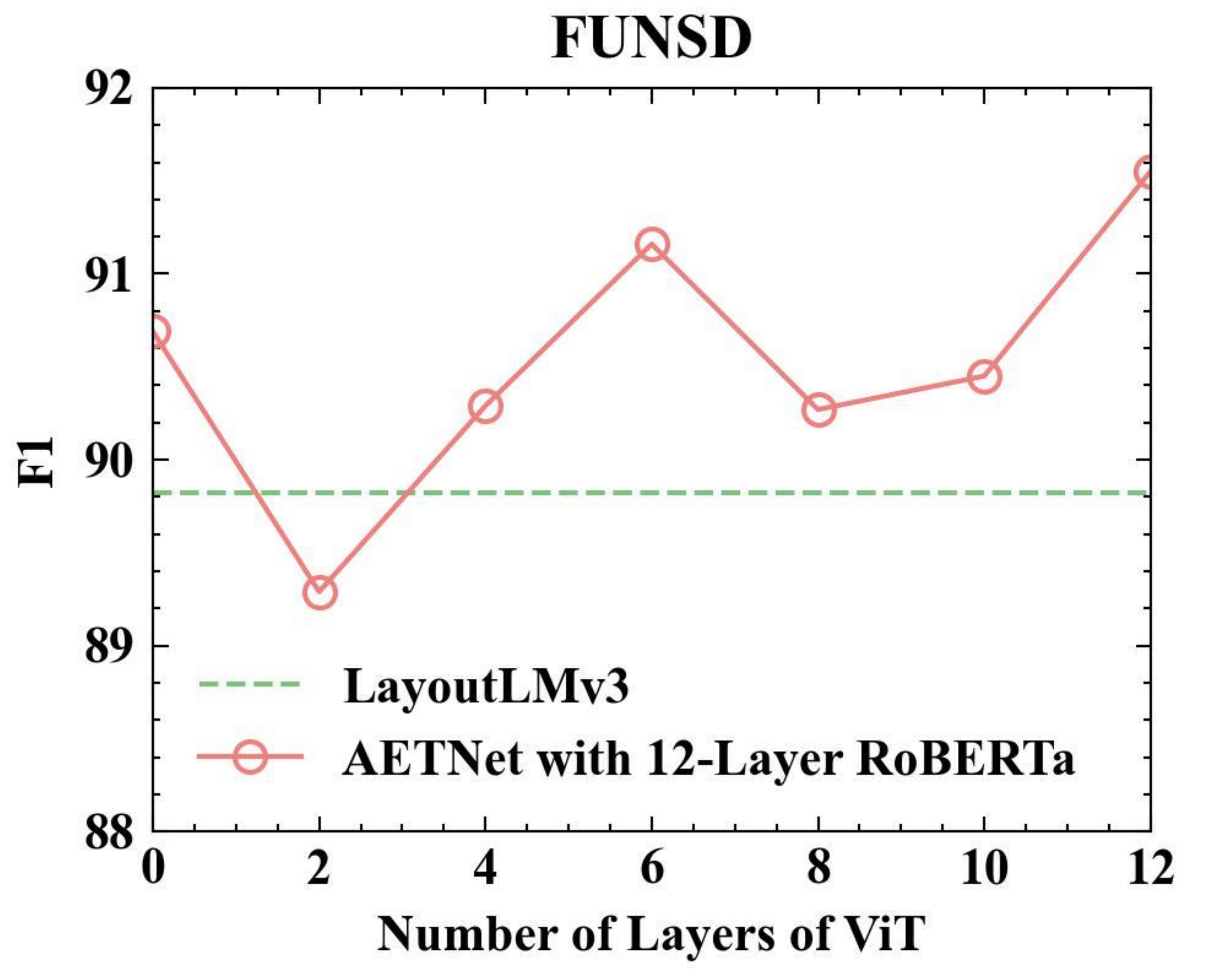}
  \caption{
  Performance (F1) comparison w.r.t. different numbers of layers of the alignment-ware image encoder (ViT$_{base}$) of AETNet$_{base}$ with the fixed 12-layer text encoder (RoBERTa$_{base}$) on the FUNSD dataset.
  }
  \label{fig:fig2}
\end{figure}

This subsection studies the impact of number of layers of the alignment-aware image encoder (ViT). First, we fix the number of layers of the alignment-aware text encoder (RoBEERTa) to be 12. We then change the number of layers of ViT from 0 to 12 with the interval of 2.  

Figure~\ref{fig:fig2} presents the results of variants of the number of layers of the ViT on the FUNSD dataset. 
We have the following observations. To our surprise, AETNet with the 0-layer ViT encoder, i.e., only using patch-level embedding layers, can achieve better performance than the one with 2-layer, 4-layer, or 8-layer ViT encoders. 
Although it can outperform LayoutLMv3 and run efficiently, it is still worse than the one with the 12-layer ViT encoder. 
The possible reason is that the model with more layers has a stronger ability to model local information, which is essential for the proposed alignment loss. 
Thus, we choose the 12-layer ViT encoder as the alignment-aware image encoder.

\subsubsection{Impact of Number of Layers of the Alignment-Aware Text Encoder}

\begin{figure}[t]
  \centering
  
  \includegraphics[width=0.86\linewidth]{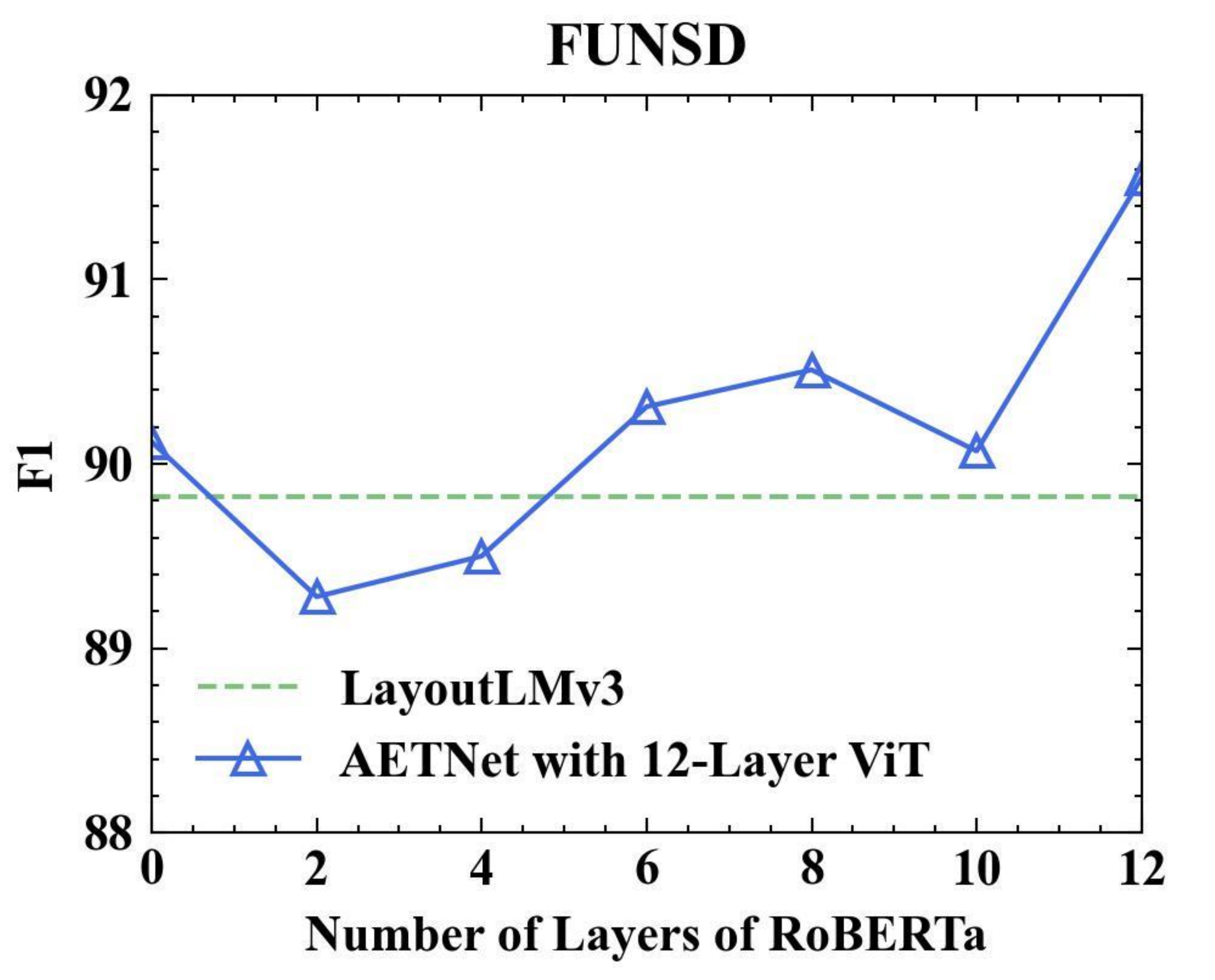}
  \caption{Performance (F1) comparison w.r.t. different numbers of layers of the alignment-ware text encoder (RoBERTa$_{base}$) of AETNet$_{base}$ with the fixed 12-layer image encoder (ViT$_{base}$) on the FUNSD dataset.}
  \label{fig:fig3}
\end{figure}

In this subsection, we study the impact of number of layers of the alignment-aware text encoder (RoBERTa). First, we fix the number of layers of the alignment-aware image encoder (ViT) to be 12. We then vary the number of layers of RoBERTa from 0 to 12 with the interval of 2.  

Figure~\ref{fig:fig3} shows the results of variants of the number of layers of RoBERTa on the FUNSD dataset. 
We observe that AETNet with the 0-layer RoBERTa encoder, i.e., only using patch-level text embedding layers, can achieve better performance than the one with 2-layer or 4-layer RoBERTa encoders. 
Beyond, it can outperform LayoutLMv3. Same as the observation of results of varying the number of ViT layers with fixed text encoder, it is still worse than the one with the 12-layer RoBERTa encoder, which achieves the best performance among all variants. 
We assume that the model with more text layers has a stronger ability to model local information hidden in the text, which is also essential for the proposed alignment loss. 
Thus, we choose the 12-layer RoBERTa encoder as the alignment-aware text encoder.

\subsection{Additional Examples of Alignment-Enriched Cases}

\begin{figure*}[t]
  \centering
  \includegraphics[width=0.84\linewidth]{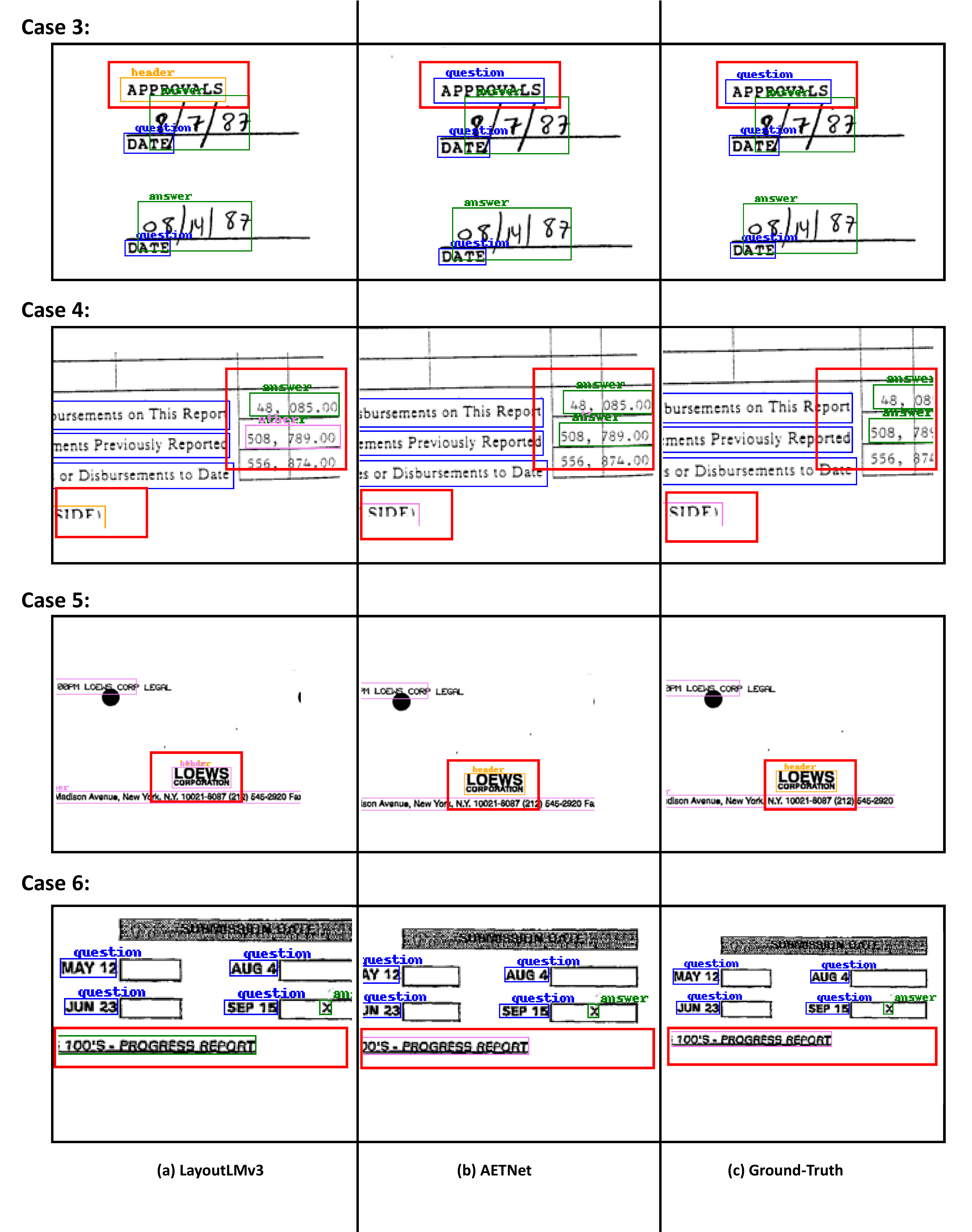}
  \caption{Visualization of case $3\sim6$ on the FUNSD dataset.}
  \label{fig:case_36 }
\vspace{-10pt}
\end{figure*}

\begin{figure*}[!ht]
  \centering
  \includegraphics[width=1.0\linewidth]{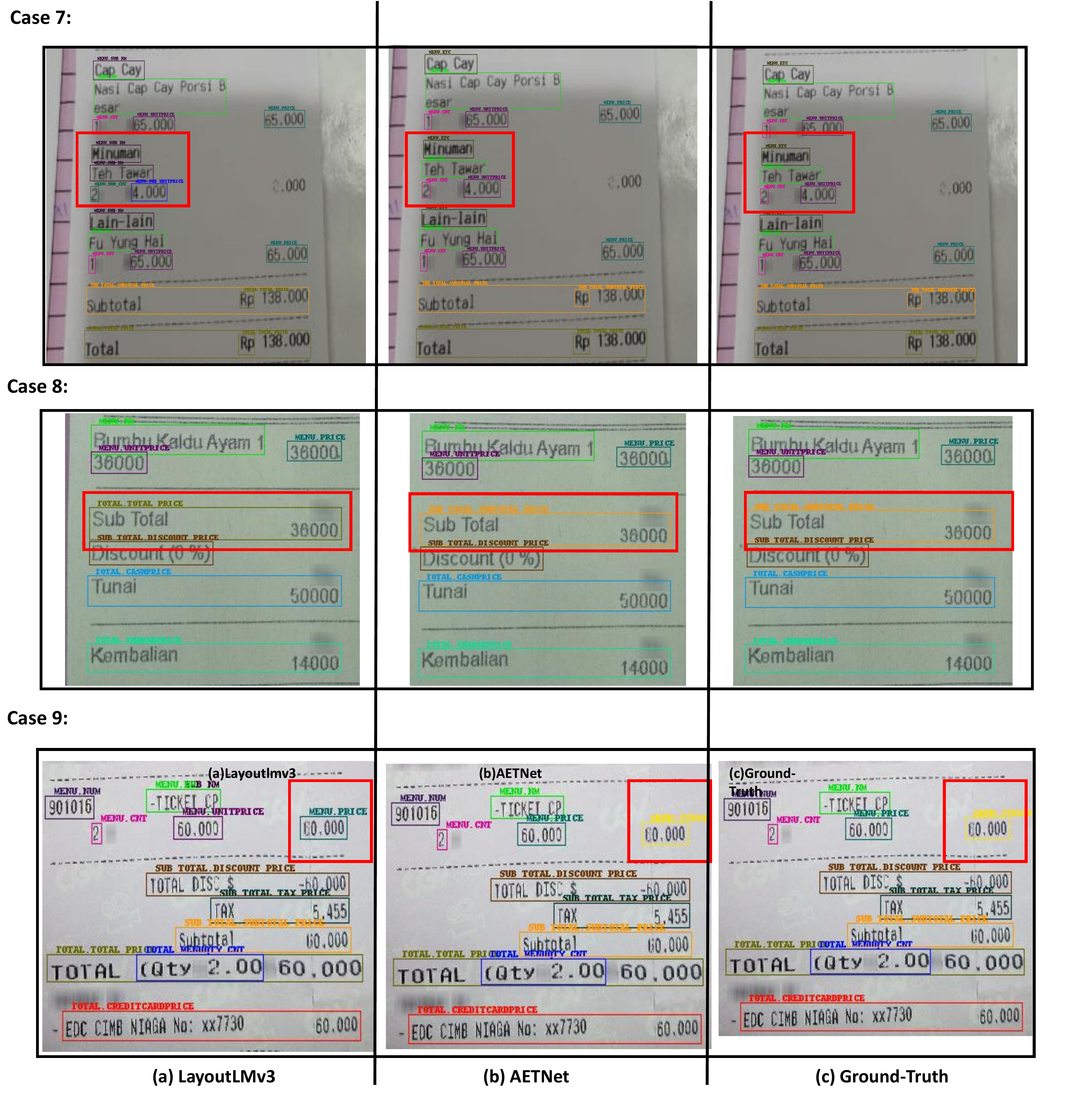}
  \caption{Visualization of case $7\sim9$ on the CORD dataset.}
  \label{fig:case_79 }
\vspace{-10pt}
\end{figure*}

In Figure~\ref{fig:case_36 } and Figure~\ref{fig:case_79 }, we show more visualization of cases on FUNSD and CORD datasets to demonstrate the strong alignment ability of our proposed AETNet to perform the document and receipt understanding.

\end{document}